\documentclass[10pt,twocolumn,letterpaper]{article}

\usepackage{wacv}
\usepackage{times}
\usepackage{epsfig}
\usepackage{graphicx}
\usepackage{amsmath}
\usepackage{amssymb}
\usepackage{url}
\usepackage{amsfonts}
\usepackage{makeidx}
\usepackage{helvet}

\wacvfinalcopy % *** Uncomment this line for the final submission
\def\wacvPaperID{274} % *** Enter the wacv Paper ID here

\ifwacvfinal\fi
\begin{document}

%%%%%%%%% TITLE
\title{One-to-many face recognition with bilinear CNNs}

% Authors at the same institution
\author{
Aruni RoyChowdhury \qquad Tsung-Yu Lin \qquad Subhransu Maji \qquad Erik Learned-Miller\\
University of Massachusetts, Amherst \\
\small{\url{{arunirc, tsungyulin, smaji, elm}@cs.umass.edu}}
}
%\author{First Author \hspace{2cm} Second Author \\
%Institution1\\
%{\tt\small firstauthor@i1.org}
%}

% Authors at different institutions
%\author{First Author \\
%Institution1\\
%{\tt\small firstauthor@i1.org}
%\and
%Second Author \\
%Institution2\\
%{\tt\small secondauthor@i2.org}
%}

\maketitle
\ifwacvfinal\thispagestyle{empty}\fi

%%%%%%%%% ABSTRACT
\begin{abstract}
The recent explosive growth in convolutional neural network (CNN)
research has produced a variety of new architectures for deep
learning. One intriguing new architecture is the bilinear CNN
(\mbox{B-CNN}), which has shown dramatic performance gains on certain
fine-grained recognition problems~\cite{lin2015bilinear}. We apply
this new CNN to the challenging new face recognition benchmark, the
IARPA Janus Benchmark~A~(IJB-A)~\cite{IJBA}. It features faces from a
large number of identities in challenging real-world
conditions. Because the face images were not identified automatically
using a computerized face detection system, it does not have the bias
inherent in such a database.  We demonstrate the performance of the
\mbox{B-CNN} model beginning from an AlexNet-style network pre-trained
on ImageNet. We then show results for fine-tuning using a
moderate-sized and public external database,
FaceScrub~\cite{ng265data}. We also present results with additional
fine-tuning on the limited training data provided by the protocol.  In
each case, the fine-tuned bilinear model shows substantial
improvements over the standard CNN. Finally, we demonstrate how a
standard CNN pre-trained on a large face database, the recently released 
VGG-Face model~\cite{Parkhi15}, can be converted into a \mbox{B-CNN} 
{\em without any additional feature training}. This \mbox{B-CNN} 
improves upon the CNN performance on the IJB-A benchmark, achieving 
89.5\% rank-1 recall.
\end{abstract}

%------------------------------------------------------------------------- 
\section{Introduction}
\label{sec:intro}
Since the introduction of the Labeled Faces in the Wild (LFW)
database~\cite{LFW}, there has been intense interest in the problem of
unconstrained face {\em verification}. In this problem, the goal is to
determine whether two face images represent the same individual or
not. From initial recognition rates in the low
seventies~\cite{nowak2007learning}, recent algorithms are achieving
rates of over 99\% using a variety of different
methods~\cite{sun2014deep,taigman2014deepface,wolf2011face}.

While face verification has been an interesting research problem, a
protocol more closely aligned with real-world applications is face
{\em identification}, also known as {\em one-to-many} or {\em 1:N}
face recognition~\cite{JanusCS2}. In this scenario, visual information
is gathered about a set of known subjects (the {\em gallery}), and at
test time, a new subject (the {\em probe}) is presented. The goal is
to determine which member of the gallery, if any, is represented by
the probe. When the tester guarantees that the probe is among the
gallery identities, this is known as {\em closed set
identification}. When the probes may include those who are not in the
gallery, the problem is referred to as {\em open set identification}.

%\begin{figure}[!htbp]
%\begin{center}
%   \includegraphics[width=0.9\linewidth]{figure/bcnn_diagram.pdf}
%\end{center}
%   \caption{\textbf{Bilinear CNN structure.} The networks A and B are chopped off at the last convolutional layer. 
%   Assuming the VGG ``M" network structure for both A and B, this gives a $512 \times W \times H$ tensor for A and B,
%   where there are 512 feature-maps arising from 512 filters; $W$ and $H$ denote the width and height of the feature maps.
%   Vectorizing along the spatial dimensions, we get $512 \times P$ matrices for each, where $P=W*H$ is the extent of the 
%   spatial dimension. The bilinear-pooling operation, $A^TB$, considers the pair-wise product of every filter and then
%   sum-pools across spatial locations.}
%\label{fig_bcnn}
%\end{figure}

Since verification accuracy rates on LFW are rapidly approaching
100\%, there has been strong demand for a new identification protocol
that can push face recognition forward again. The new \textbf{IARPA Janus
Benchmark~A~(\mbox{IJB-A})} is designed to satisfy this need.  The \mbox{IJB-A} is
presented in a CVPR paper that describes the database, gives detailed
information about proper protocols for use, and establishes initial
baseline results for the defined protocols~\cite{IJBA}.  

In this paper, we present results for \mbox{IJB-A} using the bilinear
convolutional neural network (B-CNN) of Lin
et~al.~\cite{lin2015bilinear} after adapting it to our needs and
making some minor technical changes.  In order to make use of images
from multiple perspectives, we also investigate a technique suggested by
Su~et al.~\cite{Su2015MVCNN} that pools images at the feature level,
rather than pooling classification scores. We follow the open-set 1:N
protocol and report both cumulative match characteristic (CMC) and
decision error trade-off (DET) curves, following the best practice
described in the 2014 Face Recognition Vendor Test~\cite{JanusCS2} and
suggested in the \mbox{IJB-A} paper~\cite{IJBA}.

We report results on a baseline network, a network fine-tuned with a
publicly available face database (FaceScrub) and also a network
further fine-tuned using the \mbox{IJB-A} train
set. Since \mbox{IJB-A} contains multiple images per probe, we explore
two pooling strategies as well.  We show that for the fine-tuned
networks, and for both pooling strategies, the \mbox{B-CNN}
architecture always outperforms the alternative, often by a large
margin.

Finally, we demonstrate how a pre-trained CNN can be converted into a
 \mbox{B-CNN} without any additional fine tuning of the model. 
The ``VGG-Face" CNN from Parkhi et al.~\cite{Parkhi15} was trained on a massive 
face data set which, at the time of publication, was not publicly available. 
However, the simplicity of the bilinear architecture allows the creation
of a \mbox{B-CNN} from the pre-trained CNN architecture without the need
for retraining the network. 

%This new \mbox{B-CNN} network improves over the
%performance of the VGG-Face on the \mbox{IJB-A} benchmark. This highlights
%the power of the bilinear network expansion to provide a large and rich
%set of features for classification. 

In the next subsection, we detail some
important properties of the IJB-A benchmark.

\subsection{The {IARPA} Janus benchmark {A}}
There are four key aspects of \mbox{IJB-A} that are relevant to this paper:
\begin{enumerate}
\item As stated above, it has a protocol for one-to-many classification, which is 
what we focus on.
\item It includes a wider range of poses than LFW, made possible in part by the fact that images 
were gathered by hand.
\item The data set includes both video clips and still images. 
\item Each ``observation'' of an individual is dubbed a {\em template}
  and may include any combination of still images or video clips. In
  particular, a single query or probe at test time is a template and thus typically 
  consists of multiple images of an individual.
\end{enumerate}
These aspects raise interesting questions about how to design a
recognition architecture for this problem. 

\subsection{Verification versus identification}
Identification is a problem quite different from verification.  In
{\em verification}, since the two images presented at test time are
required to be faces of individuals never seen before, there is no
opportunity to build a model for these individuals, unless it is done
at test time on the fly (an example of this is the Bayesian adaptation
of the probabilistic elastic parts model~\cite{li2013probabilistic}). Even when such
on-the-fly models are built, they can only incorporate a single image
or video to build a model of each subject in the test pair.

In {\em identification}, on the other hand, at training time the learner is
given access to a gallery of subjects with which one can learn models
of each individual.  While in some cases the gallery may contain only
a single image of a subject, in the \mbox{IJB-A}, there are typically multiple
images and video clips of each gallery subject. Thus, there is an
opportunity to learn a detailed model (either generative or
discriminative) of each subject. Depending upon the application scenario, 
it may be interesting to
consider identification systems that perform statistical learning on
the gallery, and also those that do not. In this work, adaptation to the 
gallery is a critical aspect of our performance.

\subsection{Pose variability and multiple images}
An interesting aspect of \mbox{IJB-A} is the significant pose
variability. Because the images were selected by hand, the
database does not rely on a fully automatic procedure for mining
images, such as the images of LFW, which were selected by taking the
true positives of the Viola-Jones detector.
%
%This greater pose variability suggests two important requirements for
%a successful identification system:
%\begin{itemize}
%\item The representation needs to be flexible enough to deal with a wide range of
%poses. 
%\item The architecture should have an effective way of
%  combining the information from multiple images (possibly from multiple views)
%  in a template into a single representation.
%\end{itemize}

\paragraph{Handling pose}
One way to address pose is to build 3D models~\cite{kemelmacher20113d}. In
principle, a set of gallery images can be used to generate a 3D
physical model of a head containing all of the appearance information
about the gallery identity. 

Another approach to dealing with large pose variability is to re-render
all faces from a canonical point of view, typically from a frontal pose.
While this \textit{frontalization} can be done using the above method of estimating a 3D model first,
it can also be done directly, without such 3D model estimation, as done by
\cite{taigman2014deepface}. This strategy has proven effective for obtaining high
performance on face verification benchmarks like LFW~\cite{LFW}.  It
is still not clear whether such approaches can be extended to deal
with the significantly higher variation in head poses and other
effects like occlusion that appear in \mbox{IJB-A}.

\paragraph{Averaging descriptors.}
Building a generic descriptor for each image, and simply averaging
these descriptors across multiple images in a set (i.e. a ``template")
is found to be surprisingly effective~\cite{parkhi2014compact},
yielding excellent performance in the case of verification on
LFW. However, there are clear vulnerabilities to this approach. In
particular, if a subject template contains large numbers of images from
a particular point of view, as is common in video, such averaging may
``overcount'' the data from one point of view and fail to incorporate
useful information from another point of view due to its low frequency
in a template.

Another intriguing possibility is to build a representation which,
given a particular feature representation, selects the ``best'' of
each feature across all the images in a template. This is an approach
used by the multi-view CNN of Su~et~al.~\cite{Su2015MVCNN} and
applied to the recognition of 3D objects, given multiple views of the
object. We adopt some elements of this approach and experiment
with \textit{max-pooling over feature descriptors} to get a single
representation from a collection of images.

\subsection{Face identification as fine-grained classification}
In assessing architectures for face identification, it seems natural
to consider the subarea of {\em fine-grained
  classification}. Fine-grained classification problems are
characterized by small or subtle differences among classes, and often
large appearance variability within classes. Popular fine-grained
benchmarks include such data sets as the CUB data set for bird species
identification~\cite{WelinderEtal2010}, in which each species is considered
a different class. Some pairs of species, such as different types of
gulls for example, have many features in common and can only be discriminated by
subtle differences such as the appearance of their beaks.

Face recognition can also be viewed as a fine-grained classification problem,
in which each individual represents a different class. Within-class variance is large 
and between-class variance is often subtle, making face recognition a natural member of the
fine-grained recognition class of problems.

Recently, Lin et al.~\cite{lin2015bilinear} have developed
the \textit{bilinear CNN (\mbox{B-CNN})} model specifically for
addressing fine-grained classification problems. It thus seems a
natural fit for the one-to-many face recognition (identification)
problem. By applying the \mbox{B-CNN} model to the public \mbox{IJB-A}
data set, starting from standard network structures, we are able to
substantially exceed the benchmark performance reported for the face
identification task.

In the next section, we give a brief introduction to convolutional
neural nets. Then in Section~3, we describe the bilinear
CNN of Lin et al.~\cite{lin2015bilinear}.  In Section 4, we discuss
experiments and results, and we end in Section 5 with some
conclusions and future directions.

\begin{figure}[!htbp]
\begin{center}
   \includegraphics[width=0.49\linewidth]{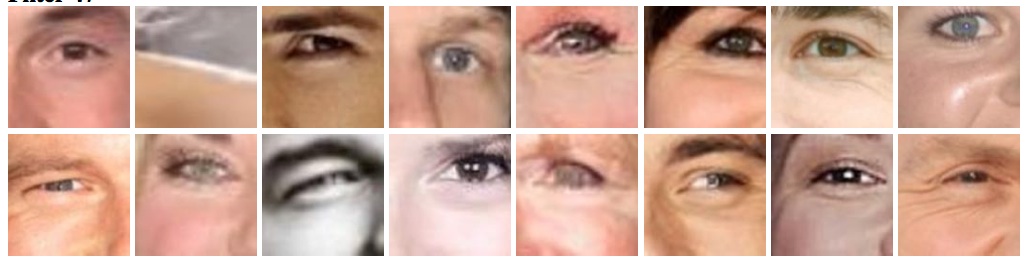}
   \includegraphics[width=0.49\linewidth]{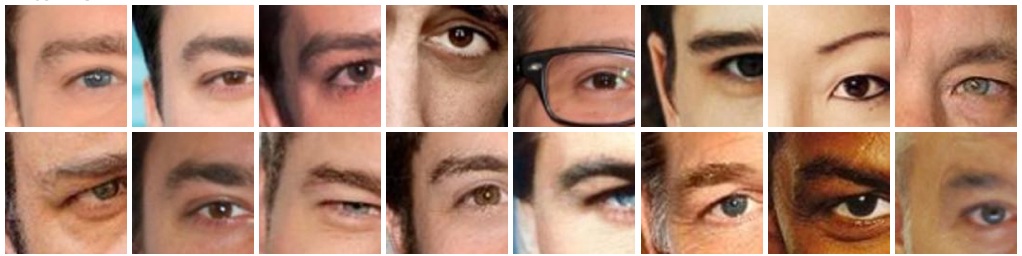}
   \includegraphics[width=0.49\linewidth]{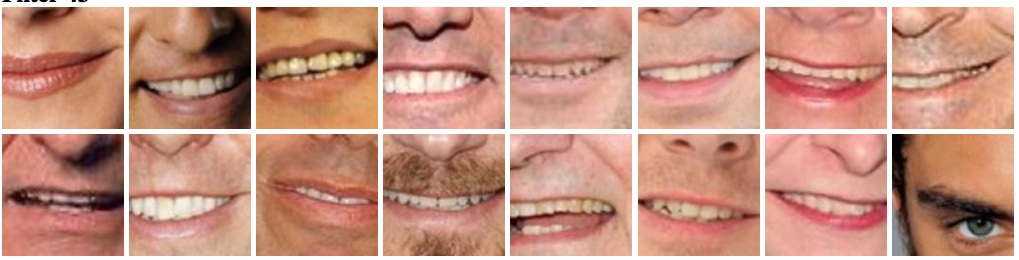}
   \includegraphics[width=0.49\linewidth]{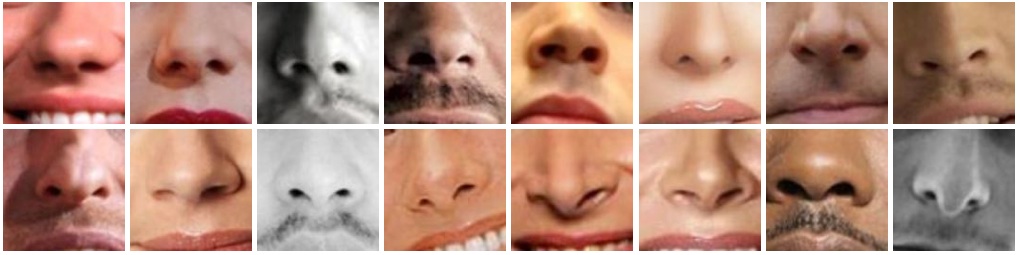} \\

   \bf{a.}

\vspace{.2in}

   \includegraphics[width=0.49\linewidth]{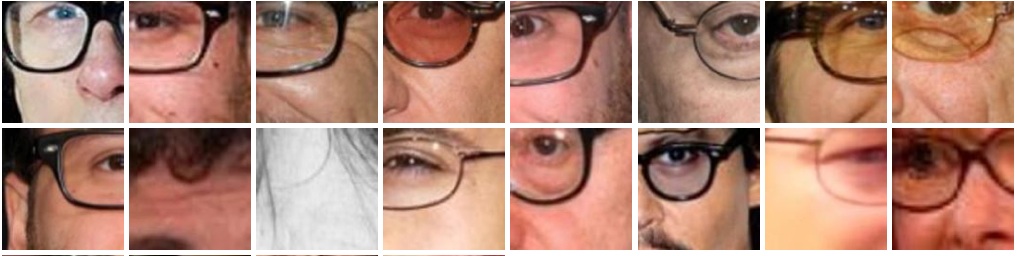}
   \includegraphics[width=0.49\linewidth]{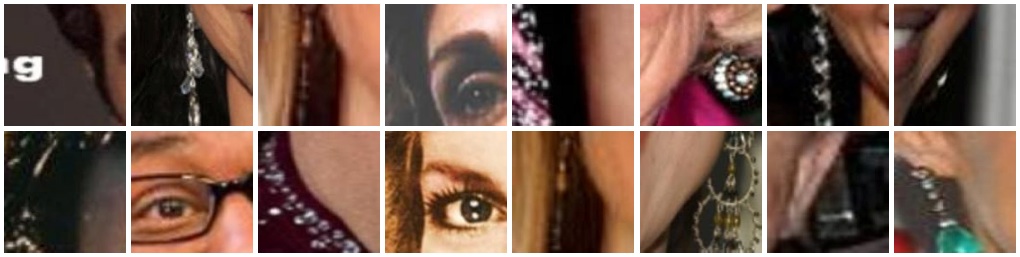}\\
   
   \bf{b.}

\vspace{.2in}

   \includegraphics[width=0.49\linewidth]{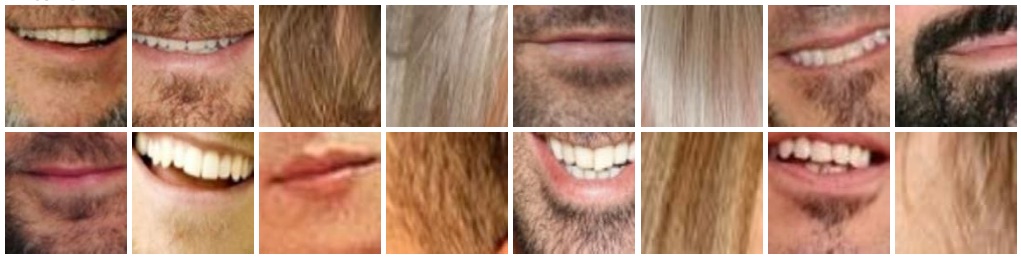}
   \includegraphics[width=0.49\linewidth]{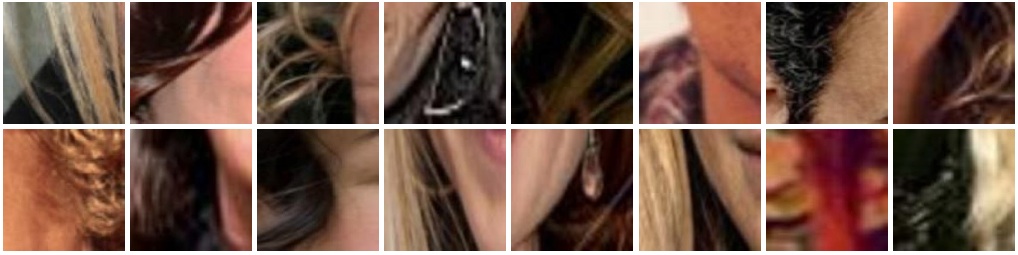}\\
   
   \bf{c.}

\vspace{.2in}

   \includegraphics[width=0.49\linewidth]{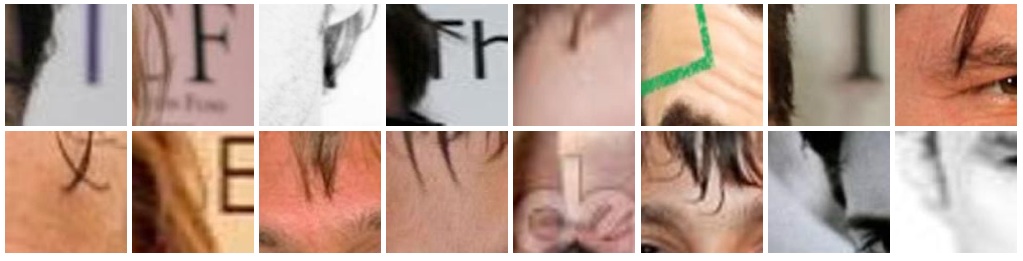}
   \includegraphics[width=0.49\linewidth]{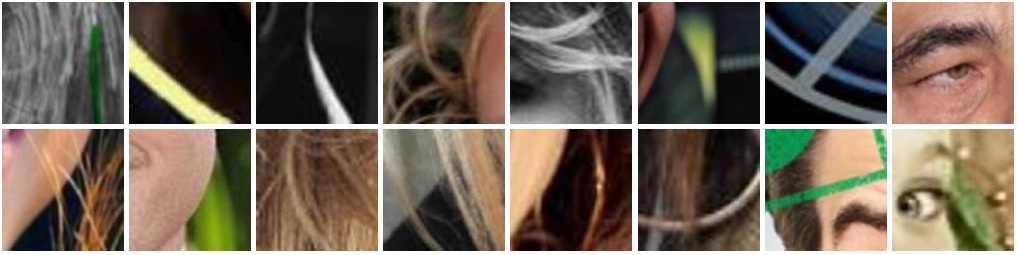}\\
   
   \bf{d.}
   
\end{center}
   \caption{Filters learned from the bilinear CNN. Each 2x8 group of
     image patches shows the top-K patches in the input images
     that gave the highest response for a particular 'conv5+relu' filter 
     of a symmetric \mbox{B-CNN}, using VGG 'M' networks and trained on 
     the FaceScrub dataset~\cite{ng265data}. 
     {\bf a.} The first 4 sets of filter
     responses show traditional facial features such as eyes,
     eyes+eyebrows, partially open mouth, and noses. 
     {\bf b.} The \mbox{B-CNN} also learns categories of features that seem to 
     be related to accessories, such as eyeglasses and jewelry.  
     {\bf c.} The features in this row seem to be correlated with hair 
     (both facial hair and hair on the top of the head). 
     {\bf d.} Finally, these features are associated with possibly 
     informative aspects of the background such as text and spurious lines.}
\label{fig_filters}
\end{figure}

\section{Introduction to CNNs}
Convolutional neural networks (CNNs) are composed of a hierarchy of units containing a convolutional, pooling (e.g. max or sum) and non-linear layer (e.g. ReLU $\max$(0, x)). In recent years deep CNNs typically consisting of the order of 10 or so such units and trained on massive labelled datasets such as ImageNet have yielded generic features that are applicable in a number of recognition tasks ranging from image classification~\cite{krizhevsky12imagenet}, object detection~\cite{girshick14rich}, semantic segmentation~\cite{hariharan2014simultaneous} to texture recognition~\cite{cimpoi15deep}. 

In the domain of fine-grained recognition, such as identifying the
breed of a dog, species of a bird, or the model of a car, these
architectures, when combined with detectors that localize various
parts of the object, have also yielded state-of-the-art
results. Without the part localization, CNNs typically don't perform
as well since there is a tremendous variation in appearance due
to different poses that instances of a category can be in. This pose
variation overwhelms the subtle differences across categories, a
phenomenon typical also in the face recognition problem. However, the
drawback of these approaches is that they require (a) manual
annotation of parts which can be time-consuming, (b) the detection of
parts which can be computationally expensive.

In contrast, models originally developed for texture recognition such as Bag-of-Visual Words (BoVM)~\cite{csurka04visual} 
and their variants such as the Fisher vector~\cite{perronnin10improving} or VLAD~\cite{jegou10aggregating}, 
have also demonstrated good results on fine-grained recognition tasks. These models don't have an explicit 
modeling of parts, nor do they require any annotations, making them easily applicable to new domains. 
Deep variants of Fisher vectors~\cite{cimpoi15deep} based on features extracted from the convolutional 
layers of a CNN trained on ImageNet~\cite{deng09imagenet} provide a better alternative to those based on 
hand-crafted features such as SIFT~\cite{lowe99object}. However, pose normalization such as ``frontalization'' 
for faces, or part detection for birds, followed by a CNN trained for fine-grained recognition outperforms 
these texture models. See for example DeepFace of Facebook~\cite{taigman2014deepface}, or pose-normalized 
CNNs for birds species identification~\cite{branson14bird, zhang2012pose}.

\section{Bilinear CNNs}
The bilinear CNN model, originally introduced by
Lin~et~al.~\cite{lin2015bilinear}, bridges the gap between the texture
models and part-based CNN models. It consists of two CNNs whose
convolutional-layer outputs are multiplied (using outer product) at
each location of the image. The resulting bilinear feature is pooled
across the image resulting in an orderless descriptor for the entire
image. This vector can be normalized to provide additional
invariances. In our experiments we follow the same protocol
as~\cite{lin2015bilinear} and perform signed square-root normalization
($\mathbf{y}\leftarrow \text{sign}(\mathbf{x})\sqrt{|\mathbf{x}|}$)
and then $\ell_2$ normalization ($\mathbf{z} \leftarrow \mathbf{y}/\|
\mathbf{y} \|$).

If one of the feature extractors was a part detector and the other
computed local features, the resulting bilinear vector can model the
representations of a part-based model. On the other hand, the bilinear
vector also resembles the computations of a Fisher vector, where the
local features are combined with the soft membership to a set of
cluster centers using an outer product (to a rough approximation, see
\cite{lin2015bilinear} for details).

A key advantage is that the bilinear CNN model can be trained using \textit{only} image 
labels without requiring ground-truth part-annotations. Since the resulting architecture 
is a directed acyclic graph (DAG), both the networks can be trained simultaneously by 
back-propagating the gradients of a task-specific loss function. This allows us to initialize 
generic networks on ImageNet and then fine-tune them on face images. Instead of having to train 
a CNN for face recognition from scratch, which would require both a search for an optimal 
architecture and a massive annotated database, we can use pre-trained networks and adapt 
them to the task of face recognition.

When using the \textit{symmetric \mbox{B-CNN}} (both the networks are identical), we can 
think of the bilinear layer being similar to the quadratic polynomial kernel often used 
with Support Vector Machines (SVMs). However, unlike a polynomial-kernel SVM, this bilinear 
feature is pooled over all locations in the image and can be trained end-to-end.

\section{Experiments}
In our experiments section, we first describe the various protocols and datasets 
used in training our models and evaluating our approach. 
In particular, the open-set protocol for face
identification and the various metrics used in the IJB-A benchmark are explained 
briefly. Next, we describe the various experimental settings for our methods. This 
includes details on data pre-processing, network architectures, fine-tuning 
procedure and using pre-trained models.

\subsection{Datasets and protocols}
\label{subsec:datasets}

\begin{figure}[!htbp]
\begin{center}
   \includegraphics[width=\linewidth]{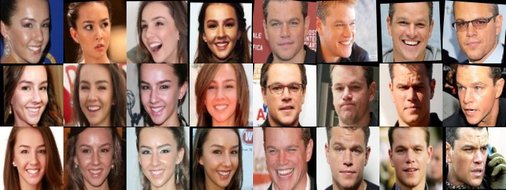}
   
   \vspace{.2in}
   
   \includegraphics[width=0.115\linewidth, height=1cm]{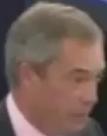}
   \includegraphics[width=0.115\linewidth, height=1cm]{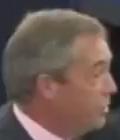}
   \includegraphics[width=0.115\linewidth, height=1cm]{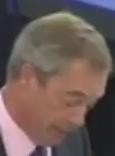}
   \includegraphics[width=0.115\linewidth, height=1cm]{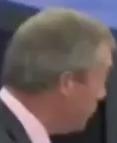}   
   \includegraphics[width=0.115\linewidth, height=1cm]{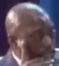}
   \includegraphics[width=0.115\linewidth, height=1cm]{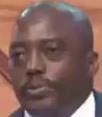}
   \includegraphics[width=0.115\linewidth, height=1cm]{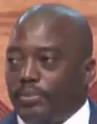}
   \includegraphics[width=0.115\linewidth, height=1cm]{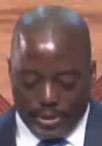}
   \\
   \includegraphics[width=0.115\linewidth, height=1cm]{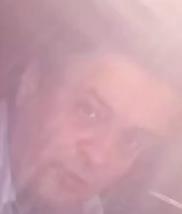}
   \includegraphics[width=0.115\linewidth, height=1cm]{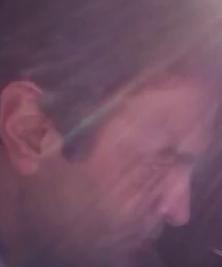}
   \includegraphics[width=0.115\linewidth, height=1cm]{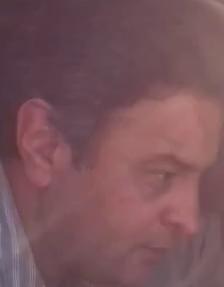}
   \includegraphics[width=0.115\linewidth, height=1cm]{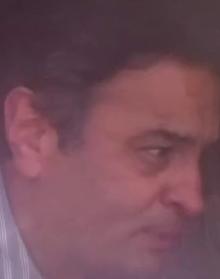}   
   \includegraphics[width=0.115\linewidth, height=1cm]{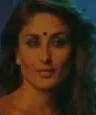}
   \includegraphics[width=0.115\linewidth, height=1cm]{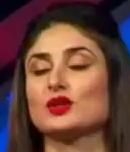}
   \includegraphics[width=0.115\linewidth, height=1cm]{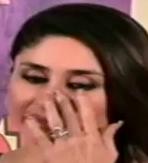}
   \includegraphics[width=0.115\linewidth, height=1cm]{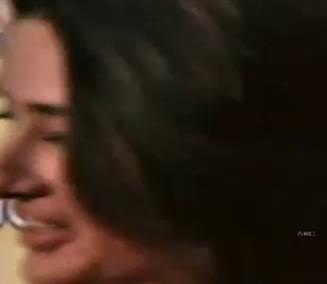}
   \\
   \includegraphics[width=0.115\linewidth, height=1cm]{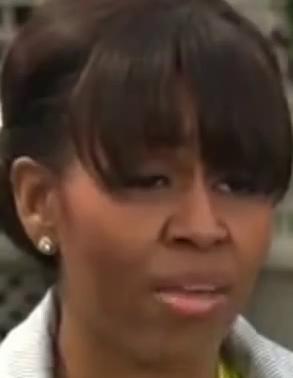}
   \includegraphics[width=0.115\linewidth, height=1cm]{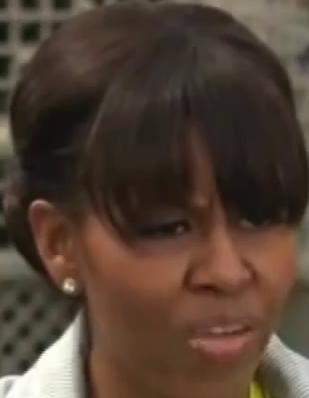}
   \includegraphics[width=0.115\linewidth, height=1cm]{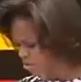}
   \includegraphics[width=0.115\linewidth, height=1cm]{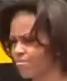}   
   \includegraphics[width=0.115\linewidth, height=1cm]{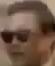}
   \includegraphics[width=0.115\linewidth, height=1cm]{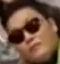}
   \includegraphics[width=0.115\linewidth, height=1cm]{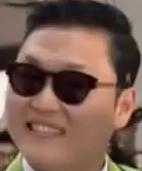}
   \includegraphics[width=0.115\linewidth, height=1cm]{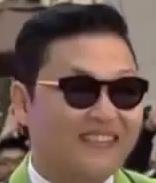}

\end{center}
   \caption{Sample images from the \textbf{FaceScrub}\protect\footnotemark \textit{(top)} and 
   \textbf{IJB-A} \textit{(bottom)} datasets.}
\label{fig_sample_img}
\end{figure}
\footnotetext{We use the sample image on the FaceScrub website \url{http://vintage.winklerbros.net/facescrub.html}}

 The \textbf{\mbox{IJB-A}} face recognition protocol~\cite{IJBA} provides
 three sets of data for each of its 10 splits.  Models can be learned
 on the {\em train set}, which contains 333 persons with varying
 number of images, including video frames, per person. The {\em
   gallery set} consists of 112 persons. The {\em probe set} is comprised of
 imagery from 167 persons, 55 of whom are not present in the gallery
 (known as ``distractors" or ``impostors").  It follows the
 \textit{open-set} protocol in its identification task.
 
 The \mbox{IJB-A} benchmark comprises both a one-to-many recognition
 task (identification) and a verification task. We focus in this paper
 on the identification task.  The details of the identification
 protocol and reporting of results can be found in the NIST report by
 Grother et al.~\cite{JanusCS2}.  To evaluate the performance of a
 system in correctly matching a probe template to its identity (from
 among the identities present in the gallery set), the \textit{Cumulative
 Match Characteristic (CMC)} curve is used. This summarizes the
 accuracy on probe templates that have a match among the gallery
 identities at various ranks.  The rank-1 and rank-5 values are
 individual points on this curve, which usually reports recall from
 ranks 1 to 100.
 
 In the open-set protocol, two particular scenarios may arise as
 follows: firstly, the ``non-match" or ``impostor" templates might be
 wrongly classified as a 
gallery identity, if the classifier score
 from the one-versus-rest SVM for that identity is above some threshold
 (false alarms). Secondly, a template that is genuinely from among the
 gallery identities may be wrongly rejected if all the SVM scores for
 it are below some threshold (misses).  The \textit{Decision Error
   Trade-off (DET)} curve plots false alarm rate or false positive
 identification rate (FPIR) and miss rate or false negative
 identification rate (FNIR) by varying this threshold, capturing both
 of the scenarios mentioned above. As specified in the \mbox{IJB-A}
 benchmark, we report FNIR at FPIR's of 0.1 and 0.01.
 
 The \textbf{FaceScrub} dataset~\cite{ng265data} is an open-access
 database of face images of actors and actresses on the web, provided
 as hyperlinks from where the actual images can be downloaded. It
 contains 530 persons with 107,818 still images in total. There are on
 average 203 images per person. In an additional experiment, we use
 this external data to first fine-tune the networks, before subsequent
 fine-tuning on the \mbox{IJB-A} train data. All overlapping identities between
 the two datasets are removed from FaceScrub before training the
 networks on it. As some of the download links provided in FaceScrub
 were broken (and after overlap removal) we finally train the networks
 on 513 identities, having a total of 89,045 images. We keep a third
 of the images in each class as validation sets and use the rest for
 training the networks.

\subsection{Methods}
\subsubsection*{Pre-processing} 
The bounding boxes provided in the \mbox{IJB-A} metadata were used to
crop out faces from the images.  The images are resized according to
the normalization parameters specified in the architecture details of
a particular network (see below). This resizing does not maintain the
aspect ratio.

\subsubsection*{Network architectures} 
As a baseline for deep models, we use the Imagenet-pretrained ``M-net" model 
from VGG's MatConvNet~\cite{vedaldi14matconvnet-convolutional}. All results 
using this network architecture are hereafter referred to as \textbf{``CNN"}. 
We consider the network outputs of the fully-connected layer after rectification, 
i.e. layer-19 (`fc7' + `relu7') to be used as the face descriptor. An input image 
is resized to $224\times224$ following the way the network had been initially trained 
on Imagenet, resulting in a 4096-dimensional feature vector.

We use a symmetric bilinear-CNN model, denoted from now on as
\textbf{\mbox{``B-CNN"}}, that has both Network A and Network B set to
the \mbox{``M-net"} model. Similar to the procedure followed
in~\cite{lin2015bilinear}, the bilinear combination is done by taking
the rectified outputs of the last convolutional layer in each network,
i.e. layer-14 (`conv5' + `relu5'). We chop off both the networks at
layer 14, add the bilinear combination layer, a layer each for
square-root and L2 normalization, and then a softmax layer for
classification.  For this architecture, the image is upsampled to be
$448\times448$, resulting in a $27\times27\times512$ output from each
network at layer-14 ($27\times27$ are the spatial dimensions of the
response map and 512 denotes the number of CNN filters at that layer).
The bilinear combination results in a $512\times512$ output, and its
vectorization (followed by the normalization layers mentioned earlier)
gives us the final face descriptor.

\subsubsection*{Network fine-tuning}
The models described in this set of experiments 
were trained initially for large-scale image classification on the
Imagenet dataset. Fine-tuning the networks for the specific task of
face recognition is expected to significantly boost performance. We
consider three different scenarios with respect to fine-tuning:

\begin{itemize}

\item \textit{no-ft:} No fine-tuning is done. We simply use the Imagenet-pretrained model without any retraining on face images as a baseline for our experiments.

%\item \textit{Train:} The networks are fine-tuned on the train set of the \mbox{IJB-A} data. We fine-tune the CNN models by replacing the last layer with a softmax, setting its learning rate to be 10 times the learning rate for the lower layers and running back-propagation with dropout regularization of 0.5 for 25 epochs. We begin fine-tuning with a learning rate of 0.001 and dividing it by 10 if the error rates does not change. The stopping time is determined when the validation error remains constant even after learning rate is changed. 
%The \mbox{B-CNN} is similarly fine-tuned for 50 epochs on the train set. It is to be noted that the identities which the networks are being trained to classify here are disjoint from the identities present in the test set. 

\item \textit{FaceScrub:} The Imagenet-pretrained network is fine-tuned on the FaceScrub dataset by replacing the 
last layer with a softmax regression and running back-propagation with dropout regularization of 0.5 for 30 epochs. 
We begin fine-tuning with a learning rate of 0.001 for the lower layers and 0.01 for the last layer and divide them
both by 10 if the validation error rate does not change. The stopping time is determined when the validation 
error remains constant even after learning rate is changed. 
The \mbox{B-CNN} is similarly fine-tuned for 70 epochs on FaceScrub data.

\item \textit{FaceScrub+Train:} The FaceScrub data provides a good
  initialization for the face identification task to the networks,
  following which we fine-tune on the \mbox{IJB-A} train set for 30
  epochs in case of the regular CNN and 50 epochs for the
  \mbox{B-CNN}. The fine-tuning on FaceScrub gives us a single model
  each for CNN and \mbox{B-CNN}. In the current setting, we take this
  network and further fine-tune it on each of the train sets provided
  in the 10 splits of \mbox{IJB-A}. This setting considers fine-tuning
  the network on images that are closer in appearance to the images it
  will be tested upon, i.e., the train set of \mbox{IJB-A}, being
  drawn from the same pool of imagery as the probe and gallery sets,
  is more similar to the test images than FaceScrub images.

\end{itemize}

\subsubsection*{Classifiers and pooling}
One-versus-rest linear SVM classifiers are trained on the gallery set
for all our experiments. We do not do any form of template-pooling at
this stage and simply consider each image or video frame of a person
as an individual sample. The weight vectors of the classifiers are
rescaled such that the median scores of positive and negative samples
are +1 and -1. Since all evaluations on this protocol are to be
reported at the template level, we have the following straightforward
strategies for pooling the media (images and video-frames) within a
template at test time:

\begin{itemize}

\item \textit{Score pooling:} We use the $\max$ operation to pool the SVM scores of all the media within a probe template. This is done after SVM scores are computed for each individual image or frame that comprises a template.

\item \textit{Feature pooling:} The $\max$ operator is applied on the features this time to get a single feature vector for a template. Thus, the SVM is run only once per probe template.

\end{itemize}

\subsubsection*{Conversion of a pre-trained CNN to a \mbox{B-CNN}}
In addition to the Imagenet-pretrained networks described above, we
also ran experiments in modifying a standard CNN, trained on a large
face database, to convert it into a \mbox{B-CNN}. We use the VGG-Face 
model from Parkhi et al.~\cite{Parkhi15}\footnote{Downloadable from 
their site \url{http://www.robots.ox.ac.uk/~vgg/software/vgg_face/}} 
where they trained the \mbox{{\tt imagenet-vgg-verydeep-16}} 
network~\cite{simonyan14very} to do face identification using a large 
data set of faces which contained no overlap in person identities 
with \mbox{IJB-A}.
This deep network architecture, pre-trained on such a large database of 
2.6 million images, produced very high results on \mbox{IJB-A} (see results
section).

We replicated this network to form a symmetric \mbox{B-CNN}, as
described in Section~4.2, and for each of the 10 splits in
\mbox{IJB-A}, trained SVMs on the output features for each gallery identity.
Note that this ``gallery training'' does not change the parameters of the
original core CNN network or the \mbox{B-CNN} produced from it.

\section{Results}

% \textit{Cumulative-Match-Characteristic} (CMC) curves for our final B-CNN model are also included (see Figure~\ref{fig_cmc}).

%We first use the Janus CS2 
%dataset to validate the various settings for our models  
%and then we evaluate the final model on the IJB-A dataset.
% As specified in the IJB-A description document~\cite{IJBA}, we 
%report the relevant parameters for the \textbf{search} paradigm of the database under the
%\textbf{closed set} protocol. 

We report two main sets of results. The first set of experiments shows that
\mbox{B-CNNs} consistently outperform standard CNNs when fine-tuned on generic
face data and also when adapted further to the specific data set at hand.
The second set of results, using the pre-trained VGG-Face network, shows that 
even in a scenario when such massive training data is not readily available 
to the end user, a \mbox{B-CNN} can be constructed, 
without further network training, that improves the final accuracy of the network.

For both sets of results, we evaluate the task of \textit{open-set}
identification on this dataset, also referred to as the ``1:N
identification task" in the IJB-A benchmark report~\cite{IJBA}. We
include the \textit{cumulative match characteristic (CMC)} curve plots
which summarizes the accuracy of the model over a range of ranks from
1 to 100.

\subsection{Fine-tuned networks}

\begin{figure*}[!htbp]
 \begin{center}
  \includegraphics[trim=4cm 8.5cm 4cm 9cm, clip=true, width=0.4\linewidth]{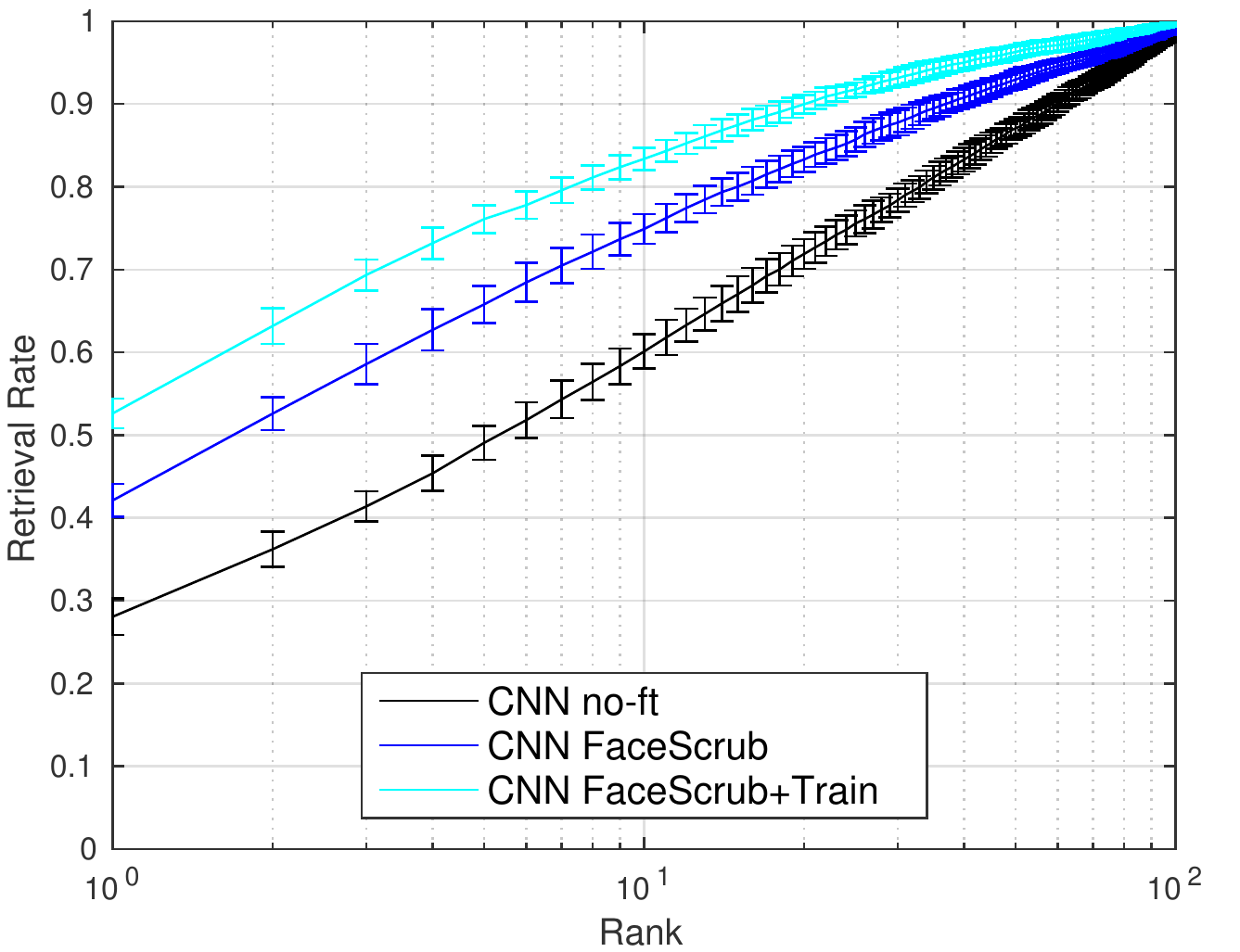}
  \hspace{0.08\linewidth}
  \includegraphics[trim=4cm 8.5cm 4cm 9cm, clip=true, width=0.4\linewidth]{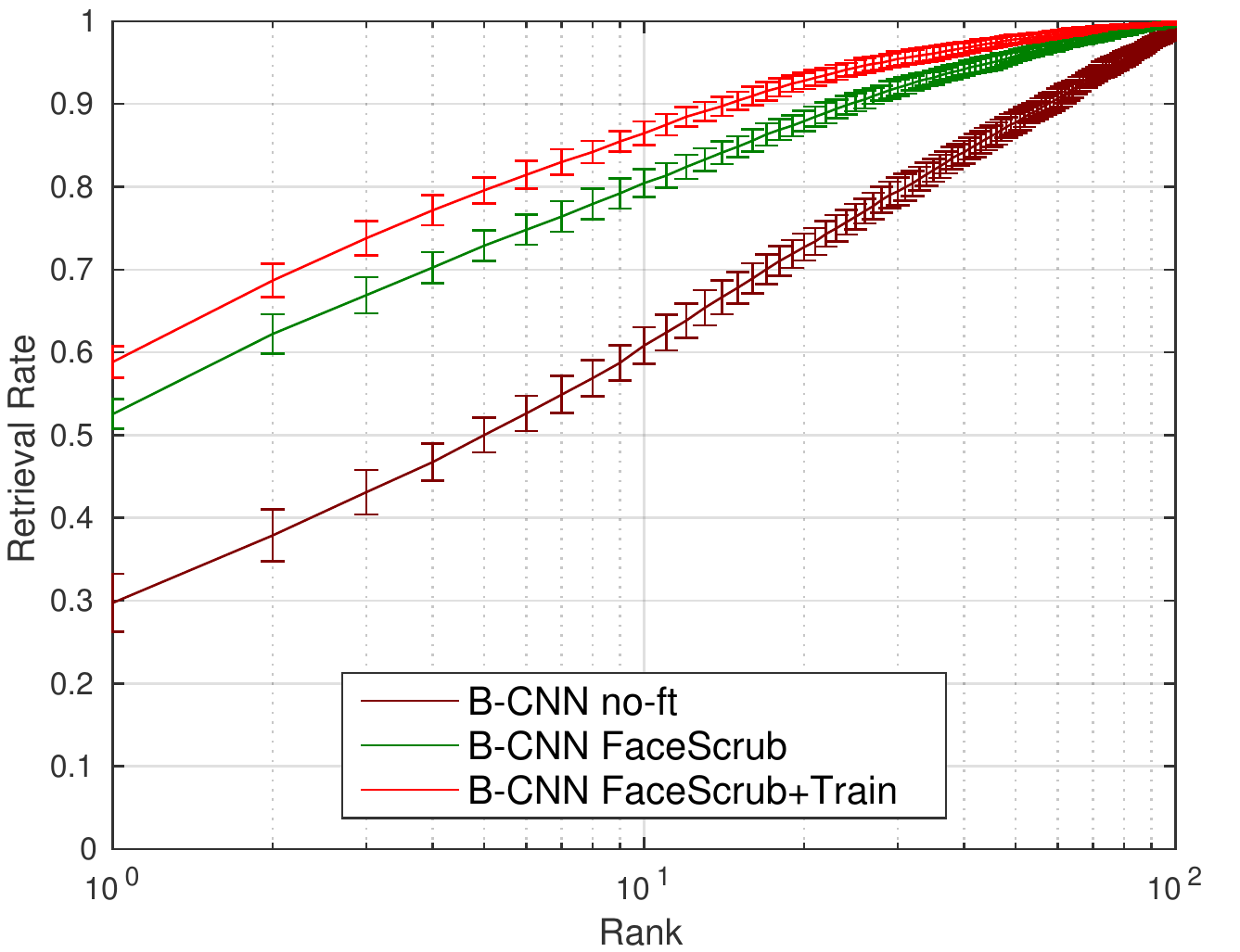}\\
 % \hspace{0.7cm}(a) \hspace{7.5cm} (b)
 \end{center}
\caption{CMC curves showing the effect of fine-tuning on the standard CNN \textit{(left)}
 and \mbox{B-CNN} \textit{(right)} models, evaluated on the IJB-A
dataset using feature pooling of templates. Both models show a large improvement when 
fine-tuned on FaceScrub when compared to the performance of Imagenet-pretrained 
networks without fine-tuning. Further fine-tuning on the IJB-A train set 
after fine-tuning on FaceScrub increases performance in 
both the models as the test and training domains are now more similar.}
\label{fig_ft}
\end{figure*}

\subsubsection*{Comparison of fine-tuning methods}
The various fine-tuning settings that are used here and the performance of the 
models at \textit{rank-1} and \textit{rank-5} 
are shown in the rows of Table~\ref{tab:comp_ijba}. Performance across ranks 
is shown in the CMC curves in Figure~\ref{fig_ft}.

Without any fine-tuning, the Imagenet-pretrained networks give low accuracy 
in both models --- 28.9\% for the standard CNN and 
31.2\% for the bilinear CNN.

Fine-tuning on FaceScrub makes the networks specific to the task of
identifying faces, and we can see highly specialized filters being
learned in the convolutional layer of the network in
Figure~\ref{fig_filters}. Consequently, the performance of both CNNs and
\mbox{B-CNNs} improves, as seen in Figure~\ref{fig_ft}. However, the 
\mbox{B-CNN} model fine-tuned on FaceScrub outperforms the CNN by almost 
10\% (52.5\% versus 44.5\%). 

\begin{table}[!htbp]
\renewcommand{\arraystretch}{1.2}
\centering
\Large
\resizebox{\columnwidth}{!}{%
\begin{tabular}{|c|cc|cc|}
\hline
\textbf{Rank-1} & \multicolumn{2}{c|}{\bf{CNN}} & \multicolumn{2}{c|}{\bf{B-CNN}} \\
                & score pooling & feature pooling & score pooling & feature pooling \\
\hline
no-ft           &  $0.289 \pm 0.028$ & $0.281 \pm 0.023$ &  $0.312 \pm 0.029$  &  $0.297 \pm 0.037$\\
FaceScrub       &  $0.445 \pm 0.020$ & $0.421 \pm 0.021$ &  $0.521 \pm 0.021$  &  $0.525 \pm 0.019$\\
FaceScrub+Train &  $0.536 \pm 0.020$ & $ 0.526 \pm 0.019$ &  $0.579 \pm 0.014$   & $\mathbf{0.588 \pm 0.020}$ \\
\hline
\textbf{Rank-5} &  &  &     &  \\
\hline
no-ft           &  $0.519 \pm 0.026$ & $0.490 \pm 0.022$ &  $0.517 \pm 0.029$  &  $0.500 \pm 0.022$\\
FaceScrub       &  $0.684 \pm 0.024$ & $0.658 \pm 0.024$ &  $0.732 \pm 0.017$  &  $0.729 \pm 0.019$ \\
FaceScrub+Train &  $0.778 \pm 0.018$ & $0.761 \pm 0.018$ & $\mathbf{0.797 \pm 0.018}$ & $0.796 \pm 0.017$ \\
\hline
\end{tabular}
}
\vspace{0.05in}
\caption{ Showing rank-1 and rank-5 retrieval rates of CNN and \mbox{B-CNN} on the \mbox{IJB-A} 1:N identification task.
		  }
\label{tab:comp_ijba}
\end{table}

 The final fine-tuning setting takes the FaceScrub trained networks
 and further fine-tunes them on the train set of \mbox{IJB-A}.
 Unsurprisingly, using training data that is very close to the
 distribution of the test data improves performance across ranks, as
 is shown in Figure~\ref{fig_ft}.  The CNN performance at rank-1
 improves from 44.5\% to 53.6\%, while the \mbox{B-CNN} performance
 rises from 52.5\% to 58.8\%.  The \mbox{IJB-A} train set images are very
 similar to the type of images in the gallery and probe test sets.
 This effect is also more prominent given that the original training
 dataset, FaceScrub, is not very extensive (530 celebrities, 107,818
 still images) compared to the large degree of variations possible in
 faces and may not cover the type of images (low resolution, motion
 blurs, video frames) seen in \mbox{IJB-A} (see the sample images in Figure \ref{fig_sample_img}).  
 However, even without depending upon the \mbox{IJB-A} train set for further 
 fine-tuning, the performance of \mbox{B-CNN} using only the single external 
 FaceScrub dataset substantially improves over the reported baselines 
 on \mbox{IJB-A} (52.5\% versus 44.3\% as shown in Table~\ref{tab:acc_ijba}).
 
\begin{figure*}[!htbp]
 \begin{center}
  \includegraphics[trim=4cm 8.5cm 4cm 9cm, clip=true, width=0.4\linewidth]{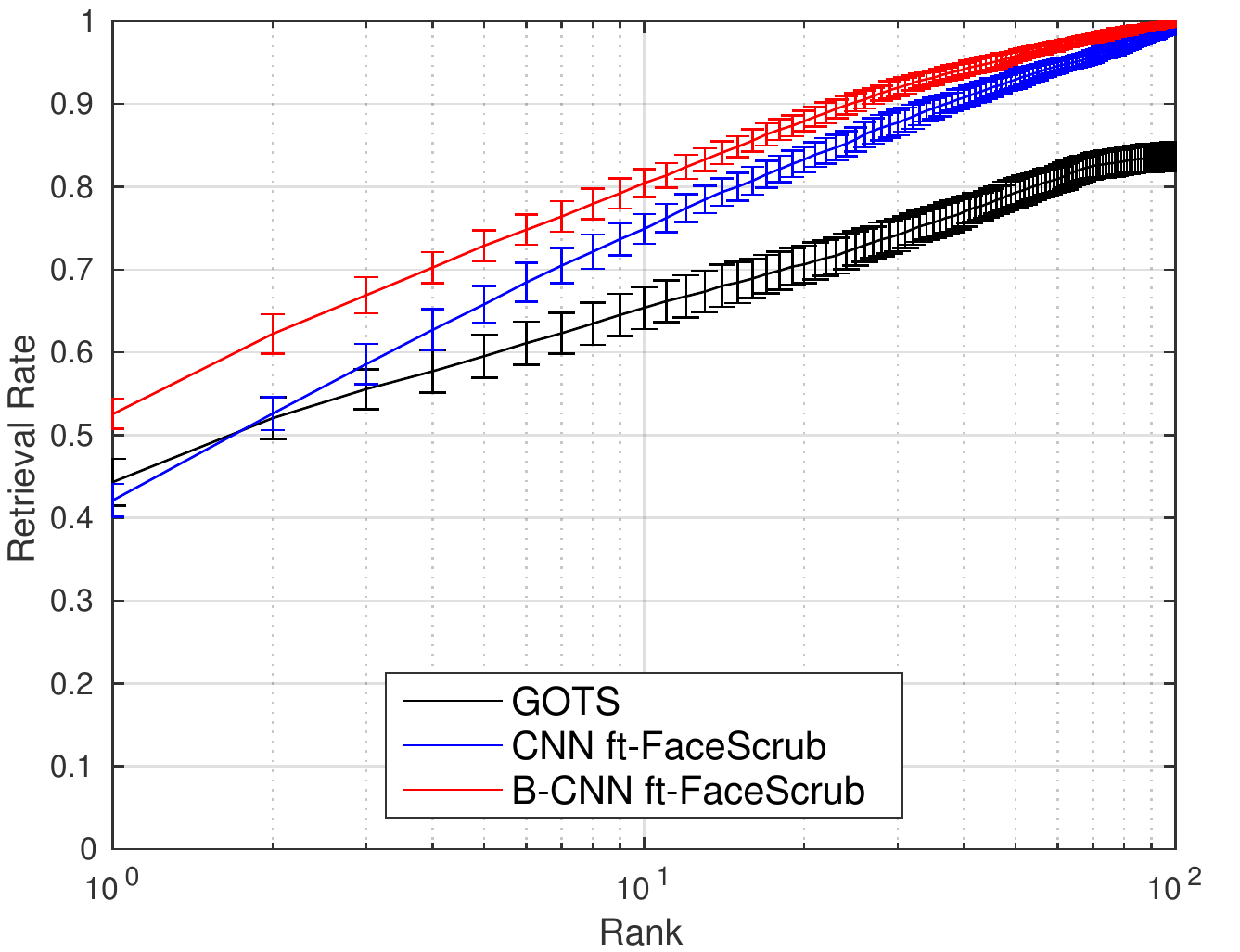}
  \hspace{0.08\linewidth}
  \includegraphics[trim=4cm 8.5cm 4cm 9cm, clip=true, width=0.4\linewidth]{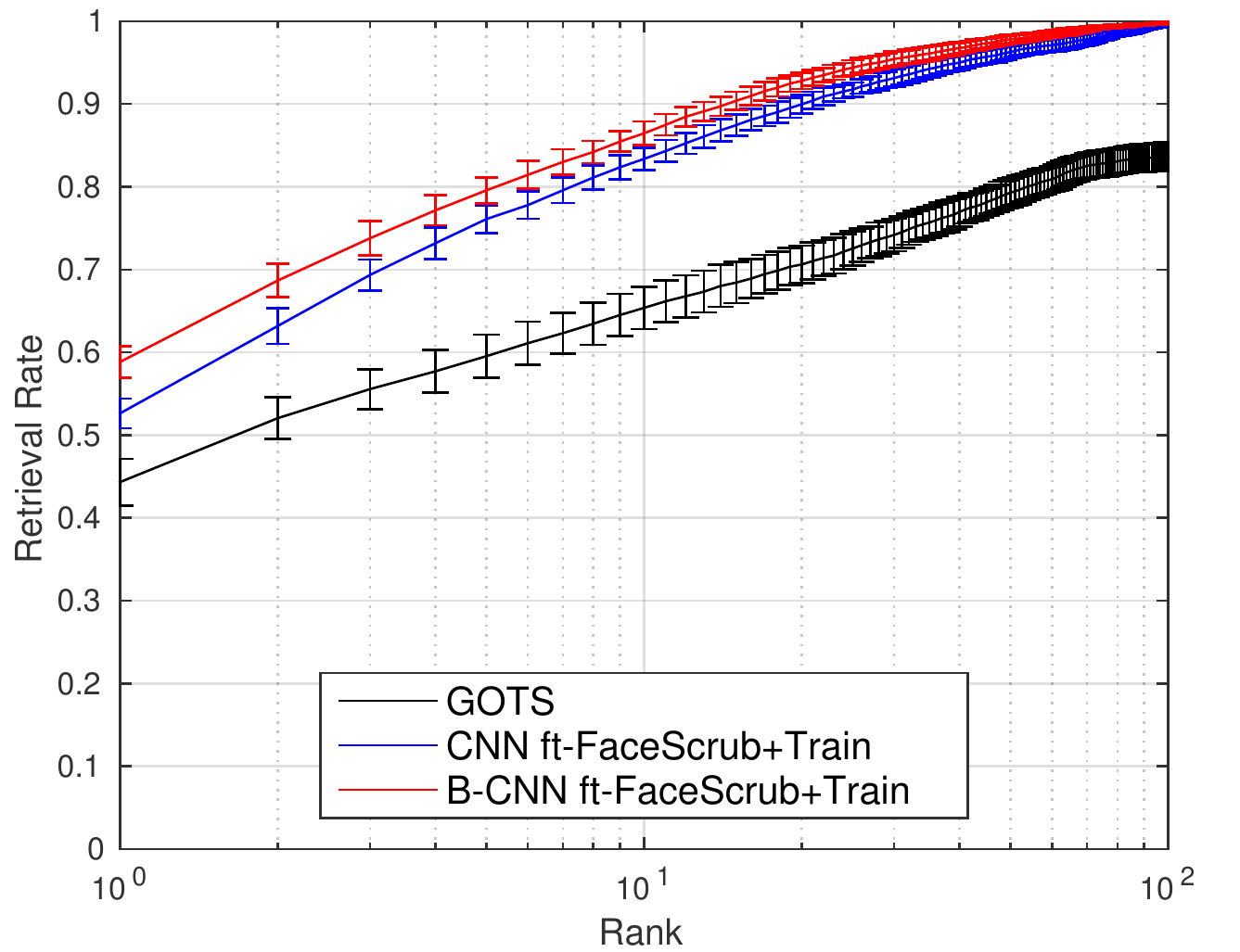}\\
 % \hspace{0.7cm}(a) \hspace{7.5cm} (b)
 \end{center}
\caption{CMC curves comparing the standard CNN and the \mbox{B-CNN} models evaluated on \mbox{IJB-A} 
 with feature-pooling of templates; 
 \textit{left:} models fine-tuned on FaceScrub; 
 \textit{right:} the FaceScrub models are fine-tuned further on \mbox{IJB-A} train set.
 The \mbox{B-CNN} does better than the CNN and also outperforms the ``GOTS" baseline method~\cite{IJBA} by a large margin.}
\label{fig_cmc}
\end{figure*}

 In Figure~\ref{fig_cmc}, we can see the consistent and large improvement in relative performance 
 by using the bilinear CNN versus the regular CNN over the entire range of ranks. The \mbox{B-CNN} also 
 outperforms the highest baseline on \mbox{IJB-A}, ``GOTS" by a large margin.
 
\vspace{-0.12in}

\subsubsection*{Comparison of template-pooling methods}
We also evaluate two pooling schemes in Table~\ref{tab:comp_ijba} at score and feature level using rank-1 and rank-5 retrieval rates. 
For the CNN network, the two pooling schemes achieve almost the same accuracy, while feature pooling either outperforms score pooling 
or has negligible difference when using fine-tuned \mbox{B-CNN} models. The reason for the preference of feature pooling on \mbox{B-CNN} 
is that the model learns good semantic part filters by fine-tuning (see Figure~\ref{fig_filters}) and feature pooling combines the part 
detections which might be invisible from a particular viewpoint by taking the maximum part detection response across views when a 
sequence of faces (with different views) is provided.
The impact of feature-pooling becomes less evident at higher ranks.

 \subsubsection*{Comparison with IJB-A benchmark}
Table~\ref{tab:acc_ijba} shows performance on \mbox{IJB-A} with respect to the published benchmark. 
The best-performing \mbox{B-CNN} model, which is trained on both FaceScrub and subsequently on \mbox{IJB-A} train set data, exceeds the 
\mbox{IJB-A} baselines~\cite{IJBA} by a large margin --- 58.8\% versus 44.3\% at rank-1 and 79.6\% versus 59.6\% at rank-5, when compared 
against the highest performing GOTS (government off-the-shelf) baseline.

\begin{figure}[!htbp]
 \begin{center}
 \includegraphics[trim=4cm 8cm 4cm 9cm, clip=true, width=0.8\linewidth]{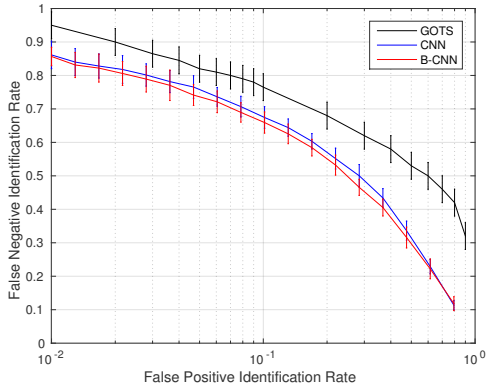}
 \end{center}
\caption{DET curve comparing the best-performing (FaceScrub+Train fine-tuned) standard CNN and the \mbox{B-CNN} models evaluated on \mbox{IJB-A}. 
 The performance of \mbox{B-CNN} is slightly better than that of CNN (lower values mean better). }
\label{fig_det}
\vspace{-0.2in}
\end{figure}

At FPIR's of 0.1 and 0.01, the FNIR of \mbox{B-CNN} versus GOTS is 65.9\% versus 76.5\% and 85.7\% versus 95.3\%, respectively. 
The DET curves for the bilinear and regular CNN models are shown in Figure~\ref{fig_det}. Our system, using 
learned bilinear features followed by the one-versus-rest linear SVMs trained on the gallery identities, is robust 
to the impostor probes without adversely affecting recognition of matching probe templates.

\begin{table}[!htbp]
\renewcommand{\arraystretch}{1.3}
\centering
\resizebox{\columnwidth}{!}{%
\begin{tabular}{|c|c|c|c|}
\hline 
\textbf{IJB-A}  & OpenBR & GOTS & \mbox{B-CNN} \\ 
\hline 
rank-1 & $0.246 \pm 0.011$ & $0.443 \pm 0.021$ & $\mathbf{0.588 \pm 0.020}$ \\ 
rank-5 & $0.375 \pm 0.008$ & $0.595 \pm 0.02$  & $\mathbf{0.796 \pm 0.017}$ \\ 
\hline 
FNIR $@$ FPIR=0.1  & $0.851 \pm 0.028$ & $0.765 \pm 0.033$ &  $\mathbf{0.659 \pm 0.032}$ \\ 
FNIR $@$ FPIR=0.01 & $0.934 \pm 0.017$ & $0.953 \pm 0.024$  & $\mathbf{0.857 \pm 0.027}$ \\ 
\hline 
\end{tabular} 
}
\vspace{0.05in}
\caption{ We compare the performance of \mbox{B-CNN} model with the baselines reported on \textbf{\mbox{IJB-A}}~\cite{IJBA}. 
The highest performing bilinear model (fine-tuned on FaceScrub and \mbox{IJB-A} train set) is considered for this comparison.}
\label{tab:acc_ijba}
\end{table}

\subsubsection*{Run-time and feature size information}
The experiments were run on an NVIDIA Tesla K40 GPU.
The \mbox{B-CNN} encoding time for a single image is roughly 0.0764
seconds. The one-versus-rest SVM training on the gallery takes around 17 seconds 
on average for a single person, using the 262,144-dimensional \mbox{B-CNN} 
features. This procedure can be easily parallelized for each person enrolled in the gallery.
The pooling methods depend largely on the number of images comprising a single
template. On average the feature pooling is 2 times faster than score
pooling at negligible difference in accuracy. Fine-tuning the bilinear CNN network 
for 70 epochs on FaceScrub took about 2 days.

% --------------------------------------------------------------------------------------------------------------------

\subsection{Pre-trained networks}

We now demonstrate that the bilinear layer can be applied out-of-the-box to further 
boost the high face recognition performance of an existing powerful pre-trained network. We evaluate 
the VGG-Face network~\cite{Parkhi15} on IJB-A as a baseline, using one-versus-rest SVM classifiers and 
no data augmentation at test time, consistent with the settings followed in our previous set of experiments. 
We compare the performance of a B-CNN built using the pre-trained VGG-Face network against the 
CNN baseline (the unmodified pre-trained network) and the method of Chen et al.~\cite{Chen_UMD_arXiv}, 
who also use a deep network trained on a large external face dataset. Table~\ref{tab:acc_vgg_ijba} 
summarizes the results, where we can see that even without fine-tuning the bilinear model, the 
\mbox{B-CNN} (at 89.5\%) outperforms both the baseline accuracy of the pre-trained CNN (89.2\%) by 
a small margin as well as the method of Chen et al. (86\%).

Without training the bilinear model, we see a only a small improvement over the regular VGG-Face CNN, 
unlike the large increase in performance observed in the previous experiments when fine-tuning of the 
bilinear model could be done. This leads us to believe that re-training the entire bilinear network formed 
out of VGG-Face on a suitably large dataset is necessary to boost the performance to a significant level 
(this is not done in our present work).

\begin{table}[!htbp]
\centering
\renewcommand{\arraystretch}{1.2}
\resizebox{\columnwidth}{!}{%
\begin{tabular}{|c|c|c|c|}
\hline 
\textbf{IJB-A}  & Chen et al.~\cite{Chen_UMD_arXiv}		& VGG-Face~\cite{Parkhi15} 	   & \mbox{B-CNN} \\ 
\hline 
rank-1 			& $0.86 \pm 0.023$ &  $0.892 \pm 0.010$ &  $\mathbf{0.895 \pm 0.011}$ \\ 
rank-5 			& $0.943 \pm 0.017$ & $0.960 \pm 0.006$  & $\mathbf{0.963 \pm 0.005}$ \\ 
\hline 
\end{tabular} 
}
\vspace{0.05in}
\caption{ We compare the performance of the pre-trained VGG-Face~\cite{Parkhi15} network with \mbox{B-CNN}, regular CNN and the
deep CNN of Chen et al.~\cite{Chen_UMD_arXiv}. Score pooling is used in both methods.}
\label{tab:acc_vgg_ijba}
\end{table}

%\begin{figure}[!htbp]
% \begin{center}
% \includegraphics[trim=1cm 8cm 1cm 9cm, clip=true, width=\linewidth]{images/CMC_IJBA_v6.pdf}
% \end{center}
%\caption{CMC curve comparing the standard CNN and the \mbox{B-CNN} models evaluated on \mbox{IJB-A} using the 
%VGG pre-trained network.}
%\label{fig_vgg_cmc}
%\end{figure}

\section{Discussion}
It is perhaps not surprising that CNN architectures that have
succeeded on other fine-grained recognition problems also do well at
face identification, both after fine-tuning and also as a simple modification 
to pre-trained models. 

%The symmetric bilinear pooling layer considers 
%all pairwise products of features and then sum-pools over the spatial locations. 
%It is possible for a deeper network with multiple fully-connected
%layers to learn something equivalent to the bilinear pooling layer. 
%However, it may take multiple such high-dimensional layers and large 
%amounts of training data to learn the mapping, whereas in performing this 
%explicitly we can get high performance at relatively low cost in terms of
%network parameters and labeled data.

There are a number of directions to continue
exploring these models:

\begin{itemize}
\item Re-training the entire model using an objective similar to the
  Multi-view CNN objective~\cite{Su2015MVCNN} in which the parameters 
  of the network are learned under the assumption that the $\max$ will be 
  taken across the multiple images in a template.
 
\item Using datasets much larger than FaceScrub, 
such as the CASIA WebFaces~\cite{yi2014learning} 
(with 10,000 identities and half-a-million images) to train the network, should further 
improve the performance. 

\item Training a very deep architecture from scratch on a sufficiently large 
face dataset, instead of fine-tuning pre-trained networks.

\end{itemize}

We believe the success of the \mbox{B-CNN} relative to non-bilinear
architectures makes them a sensible starting point for a wide variety
of continued experiments.

\section*{Acknowledgements}
This research is based upon work supported by the Office of the Director of National Intelligence (ODNI), Intelligence Advanced Research Projects Activity (IARPA), via contract number 2014-14071600010. The views and conclusions contained herein are those of the authors and should not be interpreted as necessarily representing the official policies or endorsements, either expressed or implied, of ODNI, IARPA, or the U.S. Government.  The U.S. Government is authorized to reproduce and distribute reprints for Governmental purpose notwithstanding any copyright annotation thereon.

The Tesla K40 GPU used in this research was generously donated by NVIDIA.

{\small
\bibliographystyle{ieee}
\bibliography{face_bcnn.bib}
}

\end{document}